\documentclass[10pt, a4paper]{article}

\usepackage{lrec2022} 
\usepackage{multibib}
\usepackage{graphicx}
\usepackage{multirow}
\usepackage{tabularx}
\usepackage{soul}
\usepackage{amsmath}
\usepackage{titlesec}
\titleformat{\section}{\normalfont\large\bfseries\center}{\thesection.}{1em}{}
\titleformat{\subsection}{\normalfont\SmallTitleFont\bfseries\raggedright}{\thesubsection.}{1em}{}
\titleformat{\subsubsection}{\normalfont\normalsize\bfseries\raggedright}{\thesubsubsection.}{1em}{}
\renewcommand\thesection{\arabic{section}}
\renewcommand\thesubsection{\thesection.\arabic{subsection}}
\renewcommand\thesubsubsection{\thesubsection.\arabic{subsubsection}}

\usepackage{epstopdf}
\usepackage[utf8]{inputenc}

\usepackage{hyperref}
\usepackage{xstring}

\usepackage{color}
\usepackage{float}

\title{Applying Automatic Text Summarization for Fake News Detection}

\name{Philipp Hartl, Udo Kruschwitz} 

\address{University of Regensburg, \\
         Universitätsstraße 31, 93053 Regensburg, Germany \\
         philipp1.hartl@stud.uni-regensburg.de, udo.kruschwitz@ur.de\\}

\abstract{
The distribution of \textit{fake news} is not a new but a rapidly growing problem. The shift to news consumption via social media has been one of the drivers for the spread of misleading and deliberately wrong information, as in addition to it of easy use there is rarely any veracity monitoring. Due to the harmful effects of such \textit{fake news} on society, the detection of these has become increasingly important. We present an approach to the problem that combines the power of transformer-based language models while simultaneously addressing one of their inherent problems. Our framework, CMTR-BERT, combines multiple text representations, with the goal of circumventing sequential limits and related loss of information the underlying transformer architecture typically suffers from. Additionally, it enables the incorporation of contextual information. 
Extensive experiments on two very different, publicly available datasets demonstrates that our approach is able to set new state-of-the-art performance benchmarks. Apart from the benefit of using automatic text summarization techniques we also find that the incorporation of contextual information contributes to performance gains. 
 \\ \newline  }

\begin{document}
\maketitleabstract
\section{Introduction}\label{sec:introduction}
With the rise of the Internet as the most influential information medium, the consumption of news and information has changed substantially. More recently, social media has become the primary source of information, changing this yet again \cite{shearer_news_2021}. Unfortunately, there are typically little to no checks on what information is posted and its veracity, thereby enabling the wide spread of \textit{fake news} -- intentionally and verifiably false information with the purpose of deceiving its reader \cite{allcott_social_2017}. 
The scale of the problem is such that it has become an urgent social and political issue \cite{Nakov21Automated}. The current global pandemic, for example, has even demonstrated that false information can be life-threatening \cite{marco-franco_covid-19_2021}.
Fittingly, the World Health Organization (WHO) is talking about fighting not only a pandemic, but also an infodemic \cite{hua_corona_2020}, a flood of information, including false and misleading information, which causes confusion amongst the public. Unfortunately, it is typically not a trivial task for humans to judge whether a piece of information is false or not \cite{rubin_deception_2010}.

In order to push forward the state-of-the-art (SOTA) in fake news detection we present an end-to-end deep learning approach based on the Transformer architecture \cite{vaswani_attention_2017} at the core of which we incorporate different methods to transform the original text into some condensed form. Due to architectural restrictions, transformer-based models like BERT are limited to specific input sequence lengths \cite{devlin_bert_2019}, which are shorter than many news articles \cite{souma_enhanced_2019}. To better capture the missing information, we therefore propose \textbf{CMTR-BERT} (\textbf{C}ontextual \textbf{M}ulti-\textbf{T}ext
\textbf{R}epresentations for fake news detection with \textbf{BERT}) which is an ensemble of BERT models. CMTR-BERT is particularly aimed at longer sequences and additional contextual information. The proposed model incorporates three different ways to deal with long sequences, namely a simplified \textit{hierarchical transformer} representation adopted from \newcite{pappagari_hierarchical_2019}, \textit{extractive} as well as \textit{abstractive} text summarization. Also, the model enables contextual data to be incorporated for fake news detection via additional BERT embeddings. Furthermore, the high-level architecture is language-agnostic, thereby offering plenty of future directions to reproduce our experiments in other languages. 

To the best of our knowledge, this is the first attempt at utilizing automatic text summarization to reduce text complexity for fake news classification. 


The main contributions of this work are as follows:
\begin{itemize}
    \item We propose an end-to-end deep learning framework to integrate different news and social context features for fake news detection.
    \item We use automatic text summarization techniques to circumvent information loss on long sequences.
    \item We combine different textual representations for classification.
    \item Using different benchmark datasets, we empirically investigate the influence of social context and automatic text summarization on fake news detection performance.
    \item To foster reproducibility, we make our code and models available to the community.\footnote{\url{https://github.com/phHartl/lrec_2022}}
\end{itemize}

\section{Related Work}\label{sec:background}

\subsection{Fake News Detection}\label{background:transformer}
Fake News detection systems typically adopt one of three general approaches or a combination of them \cite{sharma_combating_2019}. The most commonly used way is based on the \textit{content}, which can be either linguistic, auditory (e.g., attached voice recordings) or visual (e.g., images or videos). These approaches are often either knowledge- or style-based. The former uses methods of information retrieval to extract concrete statements, which then are automatically checked against knowledge graphs \cite{pan_content_2018} or documents retrieved from the web on the fly \cite{magdy_web-based_2010}. While these approaches are the most straightforward, wrong information and missing knowledge about recent topics limit its applicability.
The latter typically produce either certain interpretable cues \cite{vrij_criteria-based_2005,lesce_scan_1990,pennebaker_linguistic_2001} or apply a more general linguistic analysis and focus on deception or objectivity detection \cite{feng_syntactic_2012,rubin_towards_2015}. 
Linguistic approaches are however often outclassed by deep learning concepts such as LSTM architectures \cite{bahad_fake_2019} or attention-based networks \cite{bahad_fake_2019}.
Approaches which only focus on the content might miss valuable context information. Hence, \textit{context-based} solutions target secondary information such as user engagements \cite{shu_defend_2019} and dissemination networks \cite{shu_hierarchical_2020} on social platforms. These methods are based on the assumption, that there is a difference in interaction between fake and real news. This can either be done by using hand-engineered features \cite{ding_fake_2020}, propagation \cite{shu_hierarchical_2020}, temporal \cite{ferrara_what_2020} or stance analysis \cite{sobhani_detecting_2016}.  While contextual information can be useful when available, it is often not or only partially available.
\textit{Intervention-based} methods try to dynamically interpret real-time dissemination data. These are arguably the least common approaches used at the moment, because of their difficult way to evaluate \cite{sharma_combating_2019}. When used though, they try to intervene the process of fake news spreading through e.g., injecting of true news into social networks \cite{farajtabar_fake_2017} or user intervention \cite{papanastasiou_fake_2020,kim_leveraging_2018}.

\subsection{Attention-based Systems}
Recently, transformer-based approaches have led to a paradigmatic shift in NLP dominating various leaderboards ranging from question-answering\footnote{\url{https://rajpurkar.github.io/SQuAD-explorer/}} to fake news detection \cite{ding_bert-based_2020}. 
Their huge advantage comes from models like \textit{BERT} \cite{devlin_bert_2019}.
One of the main drawbacks of those, however, is the maximum sequence length each model is able to process, which comes at a maximum of 512 tokens (word pieces) for BERT. Unfortunately, fake news articles often are a longer than this value \cite{souma_enhanced_2019}. By default, BERT-based models simply truncate the text to the desired input length. This leads to the loss of potentially important information in the later parts of the input text.
To tackle this problem, \newcite{dai_transformer-xl_2019} developed Transformer-XL, which splits the input sequence in smaller chunks and injects the self-attention of the previous part into the next one as additional context with relative positional encodings. While this solves the problem of longer input sequences, it removes the bidirectional property of BERT, which is one of its major advantages and takes a lot of computing time to train.
\newcite{pappagari_hierarchical_2019} introduced hierarchical transformer representations, which is a conceptually similar approach, but built on top of BERT. 
Another possible solution is to again separate the document into different segments, but now focus on the \textsl{[CLS]} token for each segment instead. This token is designed to provide an embedding for the entire sequence (e.g a sentence). \newcite{mulyar_phenotyping_2020} use this strategy to classify clinical documents with ClinicalBERT \cite{alsentzer_publicly_2019}. 

\subsection{Automatic Text Summarization}
One of the most common and effective ways for humans to learn are summaries \cite{dunlosky_improving_2013}. Their objective is to produce a representation, which includes the main ideas of the input \cite{radev_introduction_2002}, while being also shorter than it \cite{radev_introduction_2002,tas_survey_2017} and which additionally should avoid repetitions \cite{moratanch_survey_2017}. Similar to the human concept of attention, which previously has been used successfully in language understanding in e.g. Transformers, summarizations might be able to reduce the textual scope for e.g. BERT models while also incorporating thoughts which normally would have been lost. 
Usually, summarizations are either extractive or abstractive.
\textit{Extractive} summarization focuses on extracting key phrases from the input document. These snippets are then concatenated into a summarization of desired length. The goal of \textit{abstractive} summaries is to generate a new text by paraphrasing the main concepts of the input sequence in fewer and clearer words \cite{moratanch_survey_2016}. This is a more human approach to text summarization, but also requires the ability to actually generate new text, which in itself is difficult \cite{el-kassas_automatic_2021}. 

\subsection{Datasets}
Fake News detection has often been limited by data quality and availability. 
Most datasets adapt labels directly assigned by journalists \cite{wang_liar_2017}, but there also exist approaches which condense the labelling into fewer labels \cite{shu_fakenewsnet_2020} or calculate them via a scoring system \cite{zhou_recovery_2020}. The domain varies, but tends to be of political nature \cite{silverman_hyperpartisan_2016,wang_liar_2017} or interest \cite{li_mm-covid_2020,zhou_recovery_2020}. There exist datasets which only contain short statements \cite{hanselowski_retrospective_2018} or long texts \cite{shu_fakenewsnet_2020}, based on social media text data \cite{mitra_credbank_2015,ma_detect_2017} or actual news articles \cite{norregaard_nela-gt-2018_2019}. Especially earlier datasets, only contained textual information and lacked any additional contextual information \cite{silverman_hyperpartisan_2016,wang_liar_2017}. However, due to recent research in context-based fake news detection, datasets emerged which additionally got visual, social-context and spatio-temporal information \cite{shu_fakenewsnet_2020,zhou_recovery_2020}. Most of the datasets are in English, but there also exist some in other languages \cite{vogel_fake_2019} and even multilingual ones \cite{li_mm-covid_2020}. 

\subsection{Concluding Remarks}
Our exploration of related work suggests there is a gap in applying automatic text summarization to tackle the problem of sequence limits for fake news detection. We suspect a positive influence of summarizations on classification performance due to the additional information present in the text, which normally would have been lost when using Transformers.
\section{Methodology}\label{sec:methodology}
We will now introduce 
CMTR-BERT (\textbf{C}ontextual \textbf{M}ulti-\textbf{T}ext
\textbf{R}epresentations for fake news detection with \textbf{BERT}), a BERT-based ensemble model which uses a combination of different text representations and additional context information.

\begin{figure}[ht!]
    \centering
    \includegraphics[width=\linewidth]{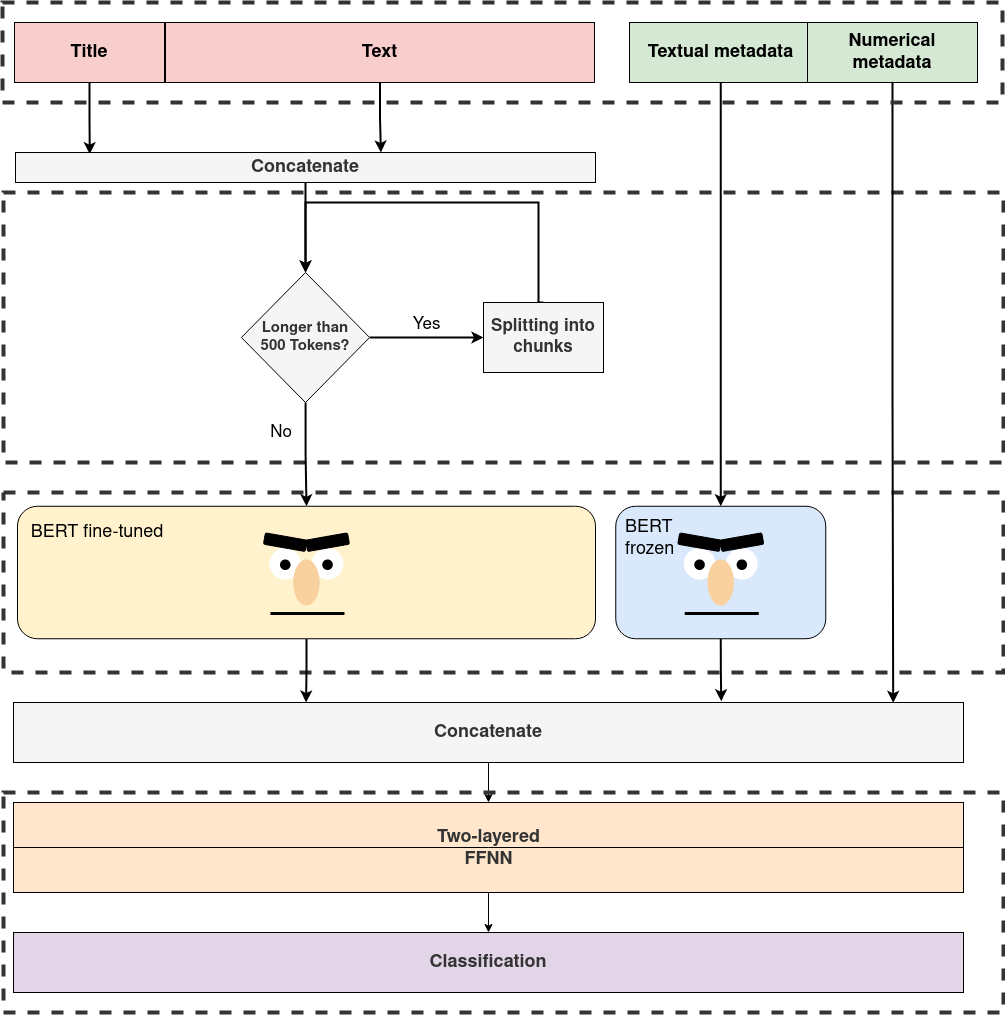}
    \caption{Model architecture (best viewed in colour)}
    \label{fig:model_architecture}
\end{figure} 

This is a hybrid approach of content- \& context-based fake news detection. It is based on the assumption of stylistic differences between fake and real information, combined with the different social reactions towards them. The basic model architecture is inspired by \newcite{ostendorff_enriching_2019}, while the idea to use automatic summarizations as additional text representations has been inspired by \newcite{li_connecting_2020}. CMTR-BERT consists of four major components for each text representation (see Figure~\ref{fig:model_architecture} from top to bottom):
\begin{enumerate}
\item Contextual feature extraction
\item Hierarchical input representation
\item A fine-tuned BERT \& frozen BERT model
\item Classification
\end{enumerate}
The following sections will explain each component in more detail.
\subsection{Contextual Feature Extraction}
To achieve the best classification performance, it is important to represent incoming information in a comprehensive way. In any text classification task, there is an input sequence of interest present. In the case of fake news, this is typically an article written by one or more agents. We consider this part to be the classification \textit{content} (marked red in Figure~\ref{fig:model_architecture}), which for news articles consists of the \textit{title} and its \textit{text}. Additional \textit{context} (marked green in Figure~\ref{fig:model_architecture}) can also be supplied when available, which can be of different modalities such as text, visual or auditorial nature. 
To be understood by machines though, content \& context information needs to be transformed into numerical values. For textual inputs, this can easily be achieved with pre-trained BERT embeddings. For other modalities, CMTR-BERT can either consider them a numerical input, e.g. when there already has been a transformation into a matrix representation or alternatively as textual input. Models like image2sentence \cite{vinyals_show_2016} provide a textual representation of inputs of other modalities.
It is very important to map the associations between \textit{content} \& \textit{context}, to get contextual content representation. We decided to represent this connection with a concatenation of \textit{content} and textual \& numerical \textit{context}.
\subsection{Hierarchical Input Representation}\label{sec:model_hierarchical}
To circumvent the problem of long sequences, we used a modified version of the hierarchical input representation (HIR) proposed by \newcite{pappagari_hierarchical_2019}. While in the original work the authors use an additional Transformer or LSTM on top of the BERT embeddings of the text chunks, we opted to not do this. Instead, we went for a concatenation of the embeddings afterwards. We did this because \newcite{mulyar_phenotyping_2020} showed that the concatenation of sequence embeddings is either on-par or better than with an additional LSTM or Transformer on top. This furthermore conceptually and computationally facilitates the model, which is a desired characteristic. We kept the splitting of the input sequence into different parts with overlap, though. 

\subsection{BERT}\label{sec:model_bert}
This component consists of two instances of 
BERT \cite{devlin_bert_2019} as seen in Figure~\ref{fig:model_architecture}. 
To differentiate learning between content and context information, we decided to use two independent BERT instances. 
Originally we considered fine-tuning them both, but it was 
not feasible with our resources. Our graphical units (GPU) simply had not enough memory for both. We decided to fine-tune with the content information and freeze the other BERT model during training.

\subsection{Classification}
The output of the aforementioned BERT models is concatenated with additional numerical context information to get the complete contextual feature representation of both content and context. The resulting representation is able to illustrate the relationship between different input sequences in a sophisticated way. To obtain a class label, this sequence is passed into a FFNN. This neural network is optimized during training and learns the differences in the aforementioned representation which separate fake from true news.
\subsection{Summarizations}\label{sec:methodlogy:summarization}
To further circumvent the loss of potentially important information, we use abstractive and extractive summarizations. Unfortunately, producing summarizations is extremely elaborate and to our knowledge there does not exist a dataset where summarizations are manually made for fake news or claim detection. Therefore, we produced such summaries automatically.
CMTR-BERT combines all three different text representations in a majority voting ensemble. Individual models (as displayed in Figure~\ref{fig:model_architecture}) are trained for the original text, abstractive and extractive summaries. Each of those then classifies the unknown input sequences, and the ensemble now decides via majority voting which class is finally assigned. 
\section{Implementation}\label{sec:implementation}
In the following section, we provide an overview of how the model described in the previous section has been implemented. We explain the used datasets, the experimental setup \& the classification problem(s). 
Everything has been implemented in Python using PyTorch \cite{paszke_pytorch_2019} and Huggingface \cite{wolf_huggingfaces_2019}. For more details, please refer to the GitHub page.
\subsection{Datasets}
We chose two datasets. FakeNewsNet  \cite{shu_fakenewsnet_2020} is a common reference collection. The other one is CT-FAN 21 \cite{shahi_ct-fan-21_2021}, published for the 2021 CLEF CheckThatLab! Fake News Detection challenge 3a \cite{nakov_clef-2021_2021}.

{\bf FakeNewsNet}:
In contrast to other datasets in the domain like LIAR \cite{wang_liar_2017} or FEVER \cite{thorne_fever_2018}, Fake\-News\-Net not only provides the actual news text but also additional social context (e.g. Twitter interactions) and spatio-temporal information (e.g. Twitter user locations).
The authors use two fact-checking websites (\textit{Politifact}
\& \textit{GossipCop}
to get relevant fake and real news. 
Both cover different domains of false information. 
The dataset is not available publicly due to legal reasons. Therefore, we used the official GitHub repository\footnote{\url{https://github.com/KaiDMML/FakeNewsNet}} to obtain our own, slightly different (e.g. due to removed articles), copy of the dataset.

{\bf CT-FAN 21}:
This dataset got four different classes to predict as defined in \newcite{shahi2021exploratory}. The distribution of each class in the provided training and test data can be seen in Table~\ref{table:ct_fan_dataset}. 
\begin{table}[ht!]
\centering
\begin{tabular}{|l|l|l|l|l|}
\hline
Dataset  & False & Partially False & True & Other \\ \hline
Training & 486   & 235             & 153    & 76  \\ \hline
Test     & 113     & 141               & 69     & 41    \\ \hline
\end{tabular}
\caption{CT FAN 21 statistics}
\label{table:ct_fan_dataset}
\end{table}

Additionally, through a data sharing agreement, it is forbidden to redistribute the dataset, identify individuals and the original entries on the fact-checking websites. Therefore, we refrained from finding this information, although it would have been useful for classification purposes as demonstrated on a similar task \cite{yuan_early_2020}. 
\subsection{Data Preparation}
Before we can use the data, there is still need for some limited pre-processing. Due to the differences in the available information in the two datasets, the preparation is not exactly the same but conceptually similar.

\begin{table}[ht!]
\centering
\begin{tabular}{|l|l|l|}
\hline
Domain     & Fake & Real  \\ \hline
Politifact & 375  & 449   \\ \hline
GossipCop  & 4761 & 14954 \\ \hline
\end{tabular}
\caption{FakeNewsNet after preprocessing}
\label{tab:fakenewsnet_adjusted}
\end{table}

{\bf FakeNewsNet}: Before doing anything else, we removed all data points which did not contain any news article information or had no file available after running the data generation script (see Table~\ref{tab:fakenewsnet_adjusted}). Afterwards, we converted the labels to numerical values. Before starting the training, we split the dataset into a training and a validation set using the common 80/20 split, which has been used before with this dataset \cite{cui_same_2019,zhou_safe_2020}. For both domains, the labels are not equally distributed, with true news being more prevalent than fake news. To circumvent this problem, we randomly oversampled the minority class during training with the imbalanced-learn package \cite{lemaitre_imbalanced-learn_2017}. Additionally, we generated \textit{abstractive} and \textit{extractive} summaries for each text once and saved them, as this took a considerable time and reduces the impact of the non-deterministic algorithms used. Before sending the text and the metadata into the model, we also tokenized and normalized the input data with the BertTokenizer.

{\bf CT-FAN 21}: As the data of CT-FAN 21 is directly available, there was no need to remove any entries. Again, we started with converting all labels to numerical values. As the four classes are not equally distributed, we applied \textit{random oversampling} as well. As there is a separate test set provided by the task organizers, we chose not to split the training data and train with the whole dataset. The generation of both summarizations, the tokenization and normalization is done the same way as for FakeNewsNet.

\subsection{Model Implementation}

\subsubsection{Content and Context Extraction}\label{subsection_features}
For the content information, we used the news article's \textit{title} and \textit{text}. For FakeNewsNet, we gathered context information in the form of the \textit{author} as well as the \textit{source URL}. For the latter, we extracted the original URL when the Wayback Machine was used to get the article. This circumvents a potential bias of deleted/removed articles being primarily fake and easily identifiable via the URL. Furthermore, we tried to incorporate different aspects of the social context features provided by FakeNewsNet. We gathered all \textit{tweet authors}, \textit{tweet texts} and the \textit{number of retweets} to have information about the interacting users, the posts and the response behaviour. We specifically did not go for any network representation as it is beyond the scope of this work.
\subsubsection{Hierarchical Input Representation}\label{sec:implementation:hierarchical}
In our hierarchical transformer variant, we split the text into \textit{overlapping parts of 500 tokens} with a \textit{stride length of 50}, as a qualitative examination resulted in these values performing the best. 
\subsubsection{BERT}\label{sec:implementation:bert}
Both instances of BERT are implemented using the \textit{bert-base-uncased} model, with 12 encoder layers and hidden dimensions of 768, which are the default values. Due to limited computational resources, we could not use a more sophisticated BERT model like RoBERTa \cite{liu_roberta_2019}.
One model learned the differences in the textual content (\textit{title} \& \textit{text}) of the news articles, while the other provided the embeddings learned by BERT for additional textual information (\textit{author}, \textit{source URL}, \textit{tweet authors} \& \textit{tweet texts}). For each of those, we concatenated the information with a view to the news article. 
\subsubsection{Classification}
After passing the BERT models, the representations are getting concatenated into a 5377-dimensional representation. This includes four parts of textual news embeddings, each of dimension 768 as a result of the hierarchical input representation, one embedding of 768 dimension of all other textual information and a single dimension to represent the number of retweets. We artificially limited the news title \& text representation to four BERT embedding blocks, as only a very small percentage (\(< 5\%\)) of the input texts is longer than that. However, a longer representation is possible if desired.
On top of that concatenation, we use a fully connected FFNN with two layers with an ReLU 
activation function. 
To calculate the loss during back propagation, we use cross entropy.
\subsubsection{Summarizations}\label{sec:implementation:summarization}
For \textit{extractive} summarization, we use the system implemented by \newcite{miller_leveraging_2019}, which already has been used before and ensures comparability \cite{li_connecting_2020}. This method first embeds the sentences using BERT, clusters them afterwards, and then finds the sentences closest to the cluster's centroids. To better resolve appearing incoherences, we furthermore use the neuralcoref library\footnote{\url{https://github.com/huggingface/neuralcoref}}. We set a summarization ratio of 0.40 empirically, which is in line with recommendation for summary length \cite{radev_introduction_2002}.

We implemented an \textit{abstractive} technique based on BART \cite{lewis_bart_2020}. This model is specifically well suited for text generation, outperforming similar ones on question-answering tasks like SQuAD \cite{lewis_bart_2020}. Because of the repetitive nature of greedy and beam search \cite{vijayakumar_diverse_2016,shao_generating_2017}, we used \textit{Top-K} \cite{fan_hierarchical_2018} and \textit{Top-p} sampling \cite{holtzman_curious_2019} for our summaries. The exact model we used is \textit{sshleifer/distilbart-cnn-12-6}\footnote{\url{https://huggingface.co/sshleifer/distilbart-cnn-12-6}}, which is a smaller BART model trained on a news summarization dataset by \newcite{hermann_teaching_2015}. In our final configuration, we used the 100 (Top-K) most likely words and a probability (Top-p) of 95\%. 
However, like BERT, BART has also a maximum sequence limit of 1024 tokens. To circumvent this problem we used the technique described in Section~\ref{sec:model_hierarchical}, however, with a length of 1000 tokens. This ensures, that all text parts are taken into consideration when producing a summarization. We also tried to get a summarization ratio of roughly 40\% for better comparability to the extractive approach. However, as both approaches are not deterministic, this cannot always be guaranteed.
\begin{table*}[t]
\centering
\resizebox{0.85\textwidth}{!}{%
\begin{tabular}{|l|l|l|l|l|l|l|l|l|l|}
\hline
Datasets                    & Metric    & SAF   & SENTI & RST   & LIWC  & HPFN           & SAFE           & dEFEND         & BERT-baseline \\ \hline
\multirow{4}{*}{Politifact} & Accuracy  & 0.691 & 0.760 & 0.796 & 0.830 & 0.843          & 0.874          & \textbf{0.904} & 0.823         \\ \cline{2-10} 
                            & Precision & -     & 0.810 & 0.821 & 0.855 & 0.835          & 0.889          & \textbf{0.902} & 0.805         \\ \cline{2-10} 
                            & Recall    & -     & 0.760 & 0.752 & 0.792 & 0.851          & 0.903          & \textbf{0.956} & 0.807         \\ \cline{2-10} 
                            & F1        & 0.706 & 0.784 & 0.785 & 0.822 & 0.843          & 0.896          & \textbf{0.928} & 0.805         \\ \hline
\multirow{4}{*}{Gossipcop}  & Accuracy  & 0.689 & 0.740 & 0.600 & 0.725 & \textbf{0.861} & 0.838          & 0.808          & 0.790         \\ \cline{2-10} 
                            & Precision & -     & 0.760 & 0.623 & 0.773 & 0.854          & \textbf{0.857} & 0.729          & 0.553         \\ \cline{2-10} 
                            & Recall    & -     & 0.740 & 0.596 & 0.637 & 0.869          & \textbf{0.937} & 0.782          & 0.693         \\ \cline{2-10} 
                            & F1        & 0.717 & 0.750 & 0.614 & 0.698 & 0.791          & \textbf{0.895} & 0.755          & 0.614         \\ \hline
\end{tabular}%
}
\caption{Baselines (best values in each row \textbf{marked}). Unless specified, values are taken from the original papers.}
\label{tab:baselines}
\end{table*}
\subsection{Experimental Setup}\label{sec:implementation:experiment}
For training, we represented each news content as a string concatenation of \textit{[CLS] + title + [SEP] + text}, where \textit{text} is either the original text or one of the two summaries produced and [CLS] is a classification token and [SEP] is a token to indicate a separation between two sentences. 
For FakeNewsNet, when applicable, we also gathered the additional metadata as described before. 
All tweet authors are concatenated together into a long string representation with a delimiter in between each. 
The \textit{tweet texts} are concatenated together into one string representation. Additionally, we removed all duplicates.
Tweets are separated by a \textit{[SEP]} token. 

For training, we use an 80/20 training/validation split for FakeNewsNet and trained with all training data for CT-FAN 21. We used the same initial random state and split for all configurations to provide comparability. We used a batch size of 8, an initial learning rate of 5e-5, a weight decay of 0.01 and three training epochs with an AdamW \cite{loshchilov_decoupled_2018} optimizer. Everything was trained on four RTX 2080 Ti with 11 GB VRAM. This process was then repeated ten times and the average values are reported. For FakeNewsNet we report \(Accuracy\), \(Precision\), \(Recall\) and \(F_{1}\), while for CT-FAN 21 we report the corresponding macro values. These metrics are typically used to measure classification performance \cite{chen_call_2018,cui_same_2019,zhou_fake_2020}.

Additionally, these trials are used as a bootstrapping method for statistical analysis. We use paired non-parametric test measures for inferential analysis, as parametric variants are not applicable here \cite{jurafsky_martin_2021}.
Hence, we use the Wilcoxon signed-rank test, Friedman test \cite{friedman_use_1937} and Nemenyi test for post-hoc analysis \cite{nemenyi_distribution-free_1963}. All statistical analysis has been implemented using scipy \cite{virtanen_scipy_2020}, pandas \cite{mckinney_data_2010} and scikit-posthocs \footnote{\url{https://scikit-posthocs.readthedocs.io/}}. We use a threshold of \(p < 0.05\) to determine whether a significant difference is present or not.

\section{Results}\label{sec:results}
We now look at the results we obtained. This includes a performance contextualization with various comparable systems found in the literature (for more detailed results of an ablation study to testify which model component contributes most to classification performance, see the GitHub project page). 
\subsection{Baseline Systems}
\begin{table*}[t]
\centering
\resizebox{0.85\textwidth}{!}{%
\begin{tabular}{|l|l|l|l|l|l|l|l|l|l|}
\hline
Dataset & Metric & BERT & SAFE & dEFEND & CMTR-BERT O & CMTR-BERT A & CMTR-BERT E & CMTR-BERT C & CMTR-BERT \\ \hline
\multirow{4}{*}{Politifact} & Accuracy & 0.924 & 0.874 & 0.904 & 0.950 & 0.948 & 0.950 & 0.912 & \textbf{0.956} \\ \cline{2-10} 
 & Precision & 0.934 & 0.889 & 0.902 & 0.953 & 0.945 & 0.941 & \textbf{0.977} & 0.958 \\ \cline{2-10} 
 & Recall & 0.897 & 0.903 & \textbf{0.956} & 0.936 & 0.943 & 0.952 & 0.827 & 0.947 \\ \cline{2-10} 
 & F1 & 0.914 & 0.896 & 0.928 & 0.944 & 0.943 & 0.946 & 0.895 & \textbf{0.952} \\ \hline
\multirow{4}{*}{Gossipcop} & Accuracy & 0.863 & 0.838 & 0.808 & 0.960 & 0.956 & 0.958 & 0.957 & \textbf{0.963} \\ \cline{2-10} 
 & Precision & 0.741 & 0.854 & 0.729 & 0.926 & 0.917 & 0.924 & 0.859 & \textbf{0.936} \\ \cline{2-10} 
 & Recall & 0.666 & 0.937 & 0.782 & 0.908 & 0.898 & 0.899 & \textbf{0.985} & 0.910 \\ \cline{2-10} 
 & F1 & 0.701 & 0.895 & 0.755 & 0.917 & 0.907 & 0.911 & 0.918 & \textbf{0.923} \\ \hline
\end{tabular}%
}
\caption{Performance of CMTR-BERT compared to SOTA. The best values in each row are \textbf{marked}.}
\label{tab:fakenewsnet:results}
\end{table*}

We compare our work against strong baselines reported in the literature. We will report results for CT-FAN 21, but will primarily focus on FakeNewsNet as that dataset contains contextual information.
\begin{itemize}
\item Two baselines use linguistic information. \textit{RST} is text-only based and extracts rhetorical features and transforms it into a tree structure. \textit{LIWC} is a linguistic cue set designed to identify psycholinguistic differences in texts \cite{pennebaker_linguistic_2001}. Both values for RST and LIWC reported here are adopted from \cite{shu_hierarchical_2020}.

\item \textit{SAF} (\textbf{S}ocial \textbf{A}rticle \textbf{F}usion) represents news content features with an encoder-decoder architecture and captures temporal social interactions with an LSTM. The results reported here are adopted from \newcite{shu_fakenewsnet_2020}.

\item \textit{SENTI} - \newcite{ding_fake_2020} trained a Naive Bayes classifier, a decision tree and a bidirectional LSTM. Their best-performing model is based on decision trees and abbreviated as \textit{SENTI} by us.

\item \textit{HPFN} (\textbf{H}ierarchical \textbf{P}ropagation \textbf{N}etwork \textbf{F}eature) combines macro- and micro-level propagation networks with additional linguistic information. Here classification happens with a range of linguistic, social and temporal features \cite{shu_hierarchical_2020}. 

\item \textit{SAFE} (\textbf{S}imilarity-\textbf{A}ware \textbf{F}ak\textbf{E} news detection) uses multi-modal (textual and visual) information \cite{zhou_safe_2020}. They specifically investigate the similarity between textual and visual information for fake news detection. They use a modified version of TEXT-CNN \cite{kim_convolutional_2014} for both textual and visual information.

\item \textit{dEFEND} (\textbf{E}xplainable \textbf{F}ak\textbf{E} \textbf{N}ews \textbf{D}etection) focuses on explainable fake news detection. It consists of a news content encoder, a user comment encoder, sentence-comment co-attention \cite{shu_defend_2019}. 

\item \textit{BERT-baseline} is a \textit{bert-base-uncased} model with default parameters and a classification layer on top. During training, only the last layer is optimized while BERT itself is frozen, apart from that the setup is identical to Section~\ref{sec:implementation:experiment}.

\end{itemize}


Detailed baseline results are shown in Table~\ref{tab:baselines}. 

\subsection{CMTR-BERT}
For a fair comparison, we built several variants of CMTR-BERT, which we compare with either content-only systems or more complex systems presented here (for detailed configurations and corresponding results please consult the GitHub repository).
We include the aforementioned BERT-baseline and a trained BERT model here, together with several CMTR-BERT variants. Additionally, there are variants of CMTR-BERT with only one text representation (\textbf{O}riginal text, \textbf{A}bstractive summaries or \textbf{E}xtractive summaries) and the complete ensemble models,  either with content data (\textbf{CMTR-BERT}) or without (\textbf{CMTR-BERT C}). 
\subsection{Performance Analysis}\label{results:performanceanalysis}
An overview of results can be seen in Table~\ref{tab:fakenewsnet:results} and  Table~\ref{tab:fakenewsnet:results:withoutmeta} (FakeNewsNet) as well as Table~\ref{tab:ctfan:results} (CT-FAN 21).

For FakeNewsNet we have the following observations:
\begin{itemize}
    \item CMTR-BERT beats state-of-the-art systems on F1 for both datasets by large margins. This indicates the effectiveness of the outlined framework on a common fake news detection dataset
    (Table~\ref{tab:fakenewsnet:results}).

    \item Interestingly, we even get the highest F1 score for GossipCop when we remove \textit{all} context features from our model, as seen in Table~\ref{tab:fakenewsnet:results:withoutmeta}.

    \item Our selection of context features seems to capture the most important aspects quite well, as CMTR-BERT C performs extraordinary well for both datasets. Being on par with SAFE or better and outclassing dEFEND by a huge margin for GossipCop.

    \item Contextual information is more important for GossipCop than it is for Politifact. Without contextual information, CMTR-BERT performs worse for GossipCop than it does for Politifact, despite being on par otherwise (see Table~\ref{tab:fakenewsnet:results:withoutmeta}).
    
    \item BERT alone performs very well for Politifact (\(F_{1} = 0.914\)) but has problems with GossipCop (\(F_{1} = 0.701\)). This performance amplifies the previous assumption, that contextual information is important for GossipCop.

    \item There seems to be little to no difference between original texts and abstractive or extractive summarizations, despite the severe reduction in textual scope.

    \item As expected, our ensemble model outperforms a single text representation in both datasets for most metrics.
\end{itemize}
\begin{table}[t]
\centering
\resizebox{\linewidth}{!}{%
\begin{tabular}{|l|l|l|l|l|l|l|}
\hline
Dataset & Metric & LIWC & \begin{tabular}[c]{@{}l@{}}CMTR-BERT O \\ w/o context\end{tabular} & \begin{tabular}[c]{@{}l@{}}CMTR-BERT A\\ w/o context\end{tabular} & \begin{tabular}[c]{@{}l@{}}CMTR-BERT E\\ w/o context\end{tabular} & \begin{tabular}[c]{@{}l@{}}CMTR-BERT\\ w/o context\end{tabular} \\ \hline
\multirow{4}{*}{Politifact} & Accuracy & 0.830 & 0.930 & 0.918 & 0.923 & \textbf{0.933} \\ \cline{2-7} 
 & Precision & 0.855 & \textbf{0.937} & 0.910 & 0.927 & 0.929 \\ \cline{2-7} 
 & Recall & 0.792 & 0.907 & 0.912 & 0.904 & \textbf{0.924} \\ \cline{2-7} 
 & F1 & 0.822 & 0.921 & 0.910 & 0.915 & \textbf{0.926} \\ \hline
\multirow{4}{*}{Gossipcop} & Accuracy & 0.725 & 0.867 & 0.858 & 0.860 & \textbf{0.869} \\ \cline{2-7} 
 & Precision & 0.773 & \textbf{0.759} & 0.737 & 0.744 & 0.772 \\ \cline{2-7} 
 & Recall & 0.637 & \textbf{0.657} & 0.642 & 0.644 & 0.650 \\ \cline{2-7} 
 & F1 & 0.698 & 0.705 & 0.685 & 0.690 & \textbf{0.706} \\ \hline
\end{tabular}%
}
\caption{Performance without additional context information. The best values in each row are \textbf{marked}}
\label{tab:fakenewsnet:results:withoutmeta}
\end{table}

As CT-FAN 21 does not provide any contextual information, we are limited to compare our different text representations here. Due to the fact, that this task is a multi-class problem, we did not use our majority ensemble here as voting draws might occur:
\begin{table*}[t]
\centering
\resizebox{1.0\textwidth}{!}{%
\begin{tabular}{|l|l|l|l|l|l|l|l|l|}
\hline
Dataset                    & Metric    & NoFake         & NLP \& IR@UNED & BERT-baseline & BERT  & \begin{tabular}[c]{@{}l@{}}CMTR-BERT O \\ w/o context\end{tabular} & \begin{tabular}[c]{@{}l@{}}CMTR-BERT A \\ w/o context\end{tabular} & \begin{tabular}[c]{@{}l@{}}CMTR-BERT E \\ w/o context\end{tabular} \\ \hline
\multirow{4}{*}{CT-FAN 21} & Accuracy  & \textbf{0.853} & 0.528          & 0.316         & 0.453 & 0.461                                                              & 0.441                                                              & 0.480                                                              \\ \cline{2-9} 
                           & Precision & -              & -              & 0.128         & 0.424 & 0.446                                                              & 0.422                                                              & \textbf{0.467}                                                     \\ \cline{2-9} 
                           & Recall    & -              & -              & 0.251         & 0.402 & 0.414                                                              & 0.384                                                              & \textbf{0.435}                                                     \\ \cline{2-9} 
                           & F1-macro  & \textbf{0.838} & 0.468          & 0.124         & 0.395 & 0.406                                                              & 0.373                                                              & 0.428                                                              \\ \hline
\end{tabular}
}
\caption{Performance of CMTR-BERT compared to submissions of 2021's CheckThatLab. The best values in each row are \textbf{marked}.}
\label{tab:ctfan:results}
\end{table*}
\begin{itemize}
    \item Our proposed model performs best here using extractive summarizations and would have been ranked in sixth place in the official runs \cite{nakov_clef-2021_2021}, indicating also a competitive performance for a multi-class problem.\footnote{The best performing system \textit{NoFake} \cite{kumari_nofake_2021} used additional context and training. 
    If we ignore system runs which used additional information in this specific task, our system ranks third, with the best one being from \newcite{martinez-rico_nlpir_2021}, which also uses Transformers.}

    \item The difference between a basic BERT model and a fine-tuned variant is a lot more pronounced here. 
    We suspect, that fine-tuning is more crucial for this problem due to the difference in data used compared to, e.g. Politifact for FakeNewsNet.

    \item For this dataset, abstractive summarizations seem to perform poor, resulting in worse performance than a normal BERT model.
\end{itemize}

\subsection{Component Analysis}\label{sec:methodoloy:ablation}
First, we investigated whether our model architecture improves the classification measurably. A one-sided Wilcoxon test between BERT and CMTR-BERT w/o context, with significant results for both Politifact (\(Z=46, p<.05\)) and GossipCop (\(Z=45, p<.05\)), indicates a performance gain. It remains, however unclear, which parts of the model contribute to this difference the most, so we conducted an additional ablation study.

The first part we investigated is the hierarchical input representation. We compared \textit{BERT} \& \textit{CMTR-BERT O w/o context}, which resulted in a significant result for CT-FAN 21 (\(Z=46, p<.05\)), but not for FakeNewsNet's Politifact (\(Z=35, p=0.25\)) and GossipCop (\(Z=38, p=0.16\)) domain.

Furthermore, we are interested in measurable differences between our textual representations, as there seems to be little difference for FakeNewsNet but considerable divergences for CT-FAN 21. For all datasets, we applied a three factored Friedman test, which resulted in significant results for GossipCop (\(\chi^2=8.60, p<0.05\)) \& CT-FAN 21 (\(\chi^2=12.60, p<0.01\)), but not for Politifact (\(\chi^2=2.92, p=0.23\)). Post-hoc tests, along with the data seen in Table~\ref{tab:fakenewsnet:results:withoutmeta} and Table~\ref{tab:ctfan:results}, suggest that models trained on abstractive summarization perform worse. However, for CT-FAN 21 prior extractive summarization improves performance.

Another major part of the model is its ensemble structure. We investigate this by comparing all aforementioned textual representations with the ensemble within a four factored Friedman test. While not significant for Politifact (\(\chi^2=5.42, p=0.14\)), the p-value is lower than before, which might suggest that with more trials an effect is measurable here. The same test on GossipCop is highly significant (\(\chi^2=1836, p<0.01\)). After running post-hoc tests, we can confirm the significant difference between CMTR-BERT O and CMTR-BERT A (\( p<0.01\)) found in the paragraph above. Additionally, there are significant effects between CMTR-BERT - CMTR-BERT A (\( p<0.01\)) and CMTR-BERT - CMTR-BERT E (\( p<0.05\)). Contextualizing this results with Table~\ref{tab:fakenewsnet:results:withoutmeta} we can deduce that CMTR-BERT performs significantly better than with abstractive or extractive summarizations alone. 

To analyse the influence of contextual data, we compare \textit{CMTR-BERT O}, \textit{CMTR-BERT A}, \textit{CMTR-BERT E} \& \textit{CMTR-BERT} with their \textit{w/o content} counterparts. The results are both highly significant for Politifact (\(Z=11, p<0.01\)) and GossipCop (\(Z=0, p<0.01\)). After further investigating the results using one-sided Wilcoxon tests for Politifact (\(Z=769, p<0.01\)) and GossipCop (\(Z=820, p<0.01\)), it becomes apparent that the model performs significantly better with context data than without. 

Additionally, we also investigated which contextual dimension has the most impact on classification performance for FakeNewsNet. For both datasets, the \textit{Source URL} seems to be the most important factor, with the \textit{Tweets} being the second most valuable.

Lastly, we investigated whether it is feasible to train on one domain of FakeNewsNet and use the classifier on the other one. Here, however, fine-tuning is actually hurting the performance and context based systems perform better. You can find the corresponding additional tables and calculations on our GitHub repository.
\section{Conclusion}\label{sec:conclusion}
We have presented an ensemble approach for fake news detection that is based on the powerful paradigm of transformer-based embeddings and utilizes text summarization as the main text transformation step before classifying a document. 

CMTR-BERT is able to achieve state-of-the-art results for a common fake news benchmark collection and provides competitive results for a second one. Our results indicate a measurable advantage of our architecture in comparison to a standard BERT model. 

Furthermore, our results emphasis the importance of context information for fake news detection once more. Not only do all context aware systems perform substantially better, it also seems feasible to not use content information at all. While each text representation individually considered does not consistently bring advantages, the combination of all three seems to be the key. It remains unclear to what extent our input transformation influences the performance, as the results here are not decisive. 

Overall, our results suggest that this is a worthwhile direction of work, and we plan to explore this further. Specifically, we are interested in using human summarizations, as arguably automatic summarization techniques are not on par with them yet and might negatively influence the system.
Ideally, there would be an additional dataset with context information as well as aforementioned corresponding summarizations. This might also be done with a small subset of, e.g. FakeNewsNet which gets manually annotated. 

We would also like to see our approach used with other datasets and different context information to get a deeper understanding into which  type of information is key for effective fake news detection. 
\section*{Acknowledgements}

To be added.

\section{Bibliographical References}\label{reference}

\bibliographystyle{lrec2022-bib}
\bibliography{literature_local}



\end{document}